\pdfoutput=1

\documentclass[a4paper,fleqn]{cas-dc}

\usepackage[numbers]{natbib}

\usepackage{float}
\restylefloat{figure}
\usepackage{subcaption}
\usepackage{algpseudocode}
\usepackage{algorithm}
\usepackage{pifont}
\usepackage{enumitem}

\usepackage{microtype}
\definecolor{caslink}{RGB}{0,86,158}
\hypersetup{citecolor=caslink, linkcolor=caslink, urlcolor=caslink}

\ExplSyntaxOn
\cs_set:Npn \__first_footerline:
{
  \group_begin:
  \small
  \sffamily
  \ifnum\theblind>0\relax
  \else
    \__short_authors: :~
  \fi
  { \rmfamily \itshape Preprint }
  \group_end:
}
\ExplSyntaxOff

\DeclareMathOperator{\norm}{norm}

\graphicspath{{figures/}}

\begin{document}
\let\WriteBookmarks\relax

\renewcommand{\floatpagefraction}{0.7}
\renewcommand{\dblfloatpagefraction}{0.7}
\setcounter{dbltopnumber}{3}

\shorttitle{}
\shortauthors{Y. L. Tun et al.}

\title[mode = title]{Beyond Easy Clients: Difficulty- and Improvement-Aware Aggregation for Federated Reinforcement Learning in Robot Navigation}

\author[1]{Ye Lin Tun}
\ead{yelintun@khu.ac.kr}

\author[1]{Chu Myaet Thwal}
\ead{chumyaet@khu.ac.kr}

\author[1]{Eui-Nam Huh}
\ead{johnhuh@khu.ac.kr}

\author[1]{Choong Seon Hong}
\cormark[1]
\ead{cshong@khu.ac.kr}

\affiliation[1]{organization={Kyung Hee University},
            country={Republic of Korea}}

\cortext[1]{Corresponding author}

\begin{abstract}
Federated reinforcement learning enables a fleet of robots to jointly learn a shared
navigation policy without exchanging raw experience. In practice, however, the robots
rarely operate in the same conditions: an open aisle is far easier to navigate than a
cluttered room, and takes less learning to master. Standard federated aggregation rules
ignore this. Uniform averaging weights every client equally, whether or not it is still
learning, and weighting by episode return is worse, since easy environments permit high
returns by construction. Both failures stem from what we call the \emph{plateau
problem}: once a client masters its environment, its updates carry little new
information even though its reported return remains high. We propose \emph{FedDIA},
which scores each client by how much its local training improves the global model it
received, amplified by the difficulty of its environment. Difficulty is inferred
from the model's own performance, without requiring any description of the
environment. Both signals are measured against the global model, so they are
comparable across clients. On a mapless navigation task with eight heterogeneous
clients, FedDIA reaches a mean success rate of $75.5\%$ across training and
held-out environments, against $68.7\%$ for the strongest federated baseline. Its
advantage holds when the models are transferred to hand-authored warehouse and
living-space layouts, or when one of the clients is defective.
\end{abstract}


\maketitle

\section{Introduction}
\label{sec:intro}

Robots increasingly operate autonomously, from warehouses and hospitals to homes
and city streets. Reinforcement learning (RL)~\cite{sutton2018reinforcement,
DBLP:journals/corr/LillicrapHPHETS15, DBLP:journals/corr/SchulmanWDRK17,
pmlr-v80-haarnoja18b} provides a way to learn the control
policies they need through interaction rather than programming them in advance. Navigation
is a common requirement in robotics: reaching a goal while avoiding obstacles. Learning
it from onboard sensors~\cite{DBLP:conf/iros/TaiPL17, DBLP:journals/arobots/XiaoLWS22} avoids
the hand-built maps and motion planners~\cite{DBLP:journals/ram/FoxBT97} that are hard to
construct for unknown or changing environments. In practice, robots rarely operate under
identical conditions: an open aisle, a cluttered room, and a narrow corridor each demand
different behavior. A fleet deployed across such conditions spans environments of
varying difficulty~\cite{DBLP:conf/aistats/JinPYWZ22}, and can improve faster by pooling
experience than by learning alone~\cite{DBLP:journals/ral/LiuWL19}. With federated
reinforcement learning (FRL)~\cite{zhuo2019fedrl, ir.2021.02}, each robot or client trains
locally and sends only its policy parameters to a server, which combines these updates
into a shared policy. Since only parameters are shared, no client reveals the layout of
its workplace.

However, not every client's update is equally valuable to the shared policy.
A naive aggregation rule such as FedAvg~\cite{DBLP:conf/aistats/McMahanMRHA17} cannot
distinguish whether a client is actively improving, and assigns equal weight to every
update. A natural alternative is to weight each client by how well its policy
performs~\cite{azar2025acwpa, DBLP:journals/access/ChuWFCZ22}, treating a higher return
as a sign of a better model. Yet this can make the problem worse. Clients in easy
environments may master the task early in training, after which their local updates
provide little new information even though their returns remain high. Weighting by return
then favors these saturated clients over clients in harder environments that are still
improving. This creates a mismatch between a client's absolute return and its actual
contribution to the shared policy. An effective aggregation rule must therefore weight
clients by how much their local training improves the shared policy, not by absolute
return. Among clients that are still improving, it should also give greater weight to
those in more difficult environments.

To meet both requirements, we propose \emph{FedDIA}, for Difficulty- and Improvement-aware
Aggregation. It measures each client against the same baseline. Before training
locally, a client evaluates how well the shared policy (global model) performs in its
environment, and evaluates its own updated policy after training. The difference between
the two evaluations measures how much the client improves the shared policy. The initial
evaluation also reveals how hard that environment is, without requiring environment
metadata. A low initial return indicates that the shared policy struggles there. Since
every client measures against the same policy, these signals are comparable across
clients. FedDIA combines the two signals into a single aggregation weight that grows
with client improvement and is amplified for harder environments. This amplification
applies only when the client is improving, and a saturated client therefore receives
no additional weight from environment difficulty.

In summary, we make the following contributions:
\begin{itemize}[leftmargin=*]
  \item We formalize the \emph{plateau problem} in heterogeneous FRL, where clients in
  easy environments saturate early and subsequent local training yields little
  improvement despite persistently high returns. Uniform averaging treats these saturated
  clients equally with those still learning, and return-based weighting assigns them even
  higher weights. This mismatch between absolute return and actual contribution motivates
  the two requirements our method targets.

  \item We propose \emph{FedDIA}, which scores each client using two signals: the
  improvement its local training brings to the shared policy and the difficulty of its
  environment. The two signals are combined so that difficulty amplifies a client's
  weight only when that client is still improving. Among those still improving, the ones
  in harder environments receive greater weight. Saturated clients are discounted
  regardless of their absolute return.

  \item On a mapless navigation task with eight heterogeneous clients, FedDIA reaches a
  mean success rate of $75.5\%$ across training and held-out environments, against
  $68.7\%$ for the strongest federated baseline.
  Its advantage carries over to two hand-authored layouts built from industrial equipment
  and household furniture, evaluated without further training. FedDIA also maintains
  its performance when the client population includes a defective one.
\end{itemize}

\section{Related Work}
\label{sec:related}

\subsection{Federated Reinforcement Learning}
\label{sec:related:frl}

Federated reinforcement learning enables multiple agents to collaboratively learn a
shared policy while keeping their interactions and data decentralized. Zhuo et al.~\cite{zhuo2019fedrl}
introduced an early FRL approach in which two agents observe the same situations through different sensory
windows and learn using a shared fusion head. Since then, FRL has been studied across a range of applications,
with Qi et al.~\cite{ir.2021.02} providing a broad survey of the field. LFRL~\cite{DBLP:journals/ral/LiuWL19} and
FGRL~\cite{tian2024fgrl} investigate shared policy learning for robot navigation. Liang et al.~\cite{liang2023fedtrl} apply FRL
to autonomous driving, while Chu et al.~\cite{DBLP:journals/access/ChuWFCZ22} study coordinated charging of plug-in
electric vehicle fleets. In healthcare, Zhou et al.~\cite{zhou2024fedorl} learn treatment policies from offline
clinical records. Tian et al.~\cite{DBLP:journals/tii/TianMH26} extend FRL to predictive-maintenance policies for manufacturing.

\subsection{Data and Environment Heterogeneity}
\label{sec:related:hetfl}

In supervised federated learning, heterogeneity typically arises from non-independent and
identically distributed (non-IID) client datasets~\cite{DBLP:conf/aistats/McMahanMRHA17,
DBLP:conf/mlsys/LiSZSTS20, DBLP:conf/icml/KarimireddyKMRS20,
DBLP:journals/corr/abs-1806-00582, DBLP:journals/ijon/ZhuXLJ21,
DBLP:conf/iclr/LiHYWZ20}.
Common approaches address it through model personalization or local regularization to reduce
client drift~\cite{DBLP:journals/corr/abs-1912-04977, DBLP:journals/spm/LiSTS20}.
In federated reinforcement learning, heterogeneity instead arises from differences in the
environments in which clients interact. Although clients may share the same task, different
environment configurations lead to different dynamics and learning conditions.
Jin et al.~\cite{DBLP:conf/aistats/JinPYWZ22} show that averaging policies across heterogeneous
environments converges to a solution whose suboptimality increases with the degree of heterogeneity.
Existing approaches to environment heterogeneity can be broadly grouped into personalization,
regularization, and aggregation. Our work follows the aggregation approach, focusing on how client
updates are weighted at the server.

\subsection{Personalization}
\label{sec:related:personalization}

The survey of Qi et al.~\cite{ir.2021.02} identifies personalized local updates as one approach to heterogeneous
environments. Jin et al.~\cite{DBLP:conf/aistats/JinPYWZ22} represent each environment with a low-dimensional
vector while personalizing the embedding layers on a shared backbone. In FedFormer~\cite{DBLP:conf/atal/HebertGPC23},
no weights are averaged; instead, each agent learns contextual relationships with other agents to incorporate
information relevant to its environment. Personalization therefore optimizes policies for individual
environments, resulting in multiple client-specific policies rather than a single deployable
policy~\cite{DBLP:journals/tnn/TanYCY23, DBLP:conf/nips/DinhTN20, DBLP:conf/iclr/ZhangSFYA21,
DBLP:conf/aaai/HuangCZWLPZ21}.
Applying the model to a new environment may require adapting its personalized component. In contrast,
we target a single shared policy that clients improve collaboratively.

\subsection{Regularization}
\label{sec:related:regularization}

Regularization constrains local training to limit how far a client's update can drift
from the global model~\cite{DBLP:conf/cvpr/LiHS21, DBLP:conf/bigcomp/TunTPPH23,
son2022comparisons}.
FedKL~\cite{DBLP:journals/jsac/XieS23} penalizes divergence from the global model's output distribution, while
FedProx~\cite{DBLP:conf/mlsys/LiSZSTS20}, a representative approach in supervised federated learning, penalizes
divergence in the model weights. Both modify the local training process by adding a regularization term to
the training loss.

\subsection{Aggregation}
\label{sec:related:aggregation}

Training a single shared policy raises the question of how client updates should be weighted
during aggregation. Uniform averaging (FedAvg~\cite{DBLP:conf/aistats/McMahanMRHA17})
weights every client equally and therefore ignores heterogeneity altogether. A natural refinement
is to weight clients by measured performance. In supervised federated learning,
FedLBW~\cite{DBLP:journals/eswa/KundrooSK26} and FedNolowe~\cite{10.1371/journal.pone.0322766}
weight clients by validation loss. In the RL setting, ACWPA~\cite{azar2025acwpa} weights
clients using a softmax over their returns, while Chu et al.~\cite{DBLP:journals/access/ChuWFCZ22}
combine reward and loss through an attention mechanism.
Because these weights derive from absolute performance, however, they conflate
environment difficulty with a client's actual contribution. A client in an easy
environment may maintain a high return and therefore a high aggregation weight even after
its local training no longer improves the shared policy.

Improvement-based weighting addresses part of this. FGRL~\cite{tian2024fgrl} weights each client
by the growth of its own performance between successive rounds. Yet this signal is
self-referential in that each client's growth is measured from its own prior model,
giving clients different starting points. The resulting values are therefore not directly comparable
across clients in different environments. Growth relative to oneself also carries no
information about environment difficulty.
Other approaches learn aggregation decisions or select clients. Using reinforcement learning,
FedAA~\cite{DBLP:conf/aaai/He0025} derives aggregation weights from model-parameter distances and
validation performance, while FAVOR~\cite{DBLP:conf/infocom/WangKNL20} selects clients based on
compressed model weights. FedPOHCS~\cite{xie2025fedpohcs} instead scores and selects clients on
their potential to improve and on how close their dynamics stay to the population average.

To our knowledge, none of these methods weights clients by improvement measured on a common
scale, or by how hard their environments are. FedDIA fills this gap by weighting each client by its
\emph{improvement over a shared baseline}, making contributions comparable across clients.
It further amplifies this weight for \emph{harder} environments, with difficulty inferred from
the same shared baseline and without requiring environment metadata.
This combination directs aggregation toward the clients that are both improving and
in the hardest environments.

\section{Problem Formulation}
\label{sec:problem}

\subsection{Federated Reinforcement Learning Setup}
\label{sec:problem:frl}

\begin{figure*}[tbp]
  \centering
  \includegraphics[width=0.9\textwidth]{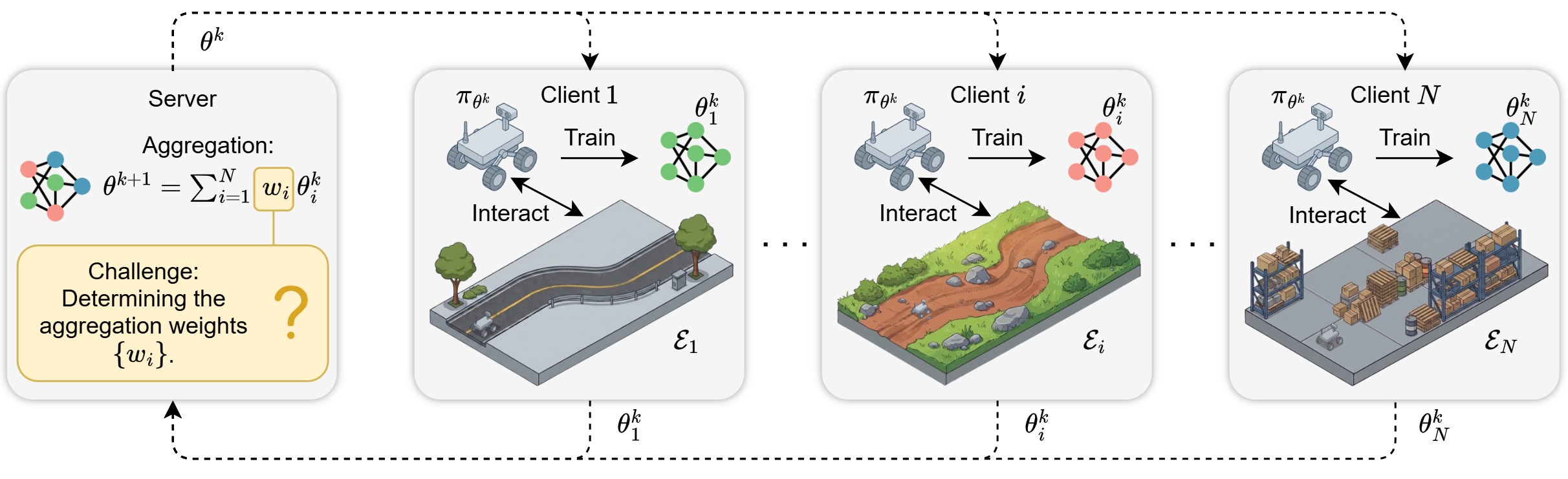}
  \caption{Federated reinforcement learning across heterogeneous environments. In each
  communication round $k$, the server broadcasts the global model $\theta^k$ to all $N$
  clients. Each client trains the policy $\pi_{\theta^k}$ in its own environment
  $\mathcal{E}_i$, and returns the updated parameters $\theta_i^k$. The
  server combines these into the next global model $\theta^{k+1}$ using aggregation
  weights $\{w_i\}$.}
  \label{fig:frl_setup}
\end{figure*}

We consider $N$ clients, where each client $i \in \{1, \ldots, N\}$ operates in its own
navigation environment $\mathcal{E}_i$.
Each client's task is modeled as a Markov decision process
(MDP)~\cite{sutton2018reinforcement}
$(\mathcal{S}, \mathcal{A}, \mathcal{P}_i, \mathcal{R}, \gamma)$,
where $\mathcal{S}$ is the state space, $\mathcal{A}$ is the continuous action space,
$\mathcal{P}_i$ is the transition function of $\mathcal{E}_i$,
$\mathcal{R} : \mathcal{S} \times \mathcal{A} \to \mathbb{R}$ is the reward function,
and $\gamma \in [0, 1)$ is the discount factor.
All clients share $\mathcal{S}$, $\mathcal{A}$, $\mathcal{R}$, and $\gamma$, and differ
only in $\mathcal{P}_i$, which reflects the layout or obstacle density of each environment.
The clients jointly train a shared policy $\pi_\theta : \mathcal{S} \to \mathcal{A}$,
parameterized by $\theta$, to maximize the expected discounted return in every
environment,
\begin{equation}
  J_i(\theta) = \mathbb{E}_{\pi_\theta, \mathcal{P}_i}
              \!\left[ \sum_{t=0}^{\infty} \gamma^t \, \mathcal{R}(s_t, a_t) \right],
  \label{eq:objective}
\end{equation}
where $s_t \in \mathcal{S}$ and $a_t \in \mathcal{A}$ denote the state and action at step $t$.
We refer to $\pi_\theta$ as the \emph{policy} and $\theta$ as the \emph{model} or,
equivalently, its \emph{parameters}.

Training is organized into communication rounds indexed by $k$, as illustrated in Figure~\ref{fig:frl_setup}.
At each round, the server broadcasts the current global parameters $\theta^k$ to all
clients.
Each client $i$ initializes from $\theta^k$, locally trains in $\mathcal{E}_i$, and sends
the resulting local parameters $\theta_i^k$ to the server.
The server aggregates the local parameters into the next global model via a weighted average:
\begin{equation}
  \theta^{k+1} = \sum_{i=1}^{N} w_i \, \theta_i^k,
  \qquad
  \sum_{i=1}^{N} w_i = 1, \quad w_i \geq 0.
  \label{eq:aggregation}
\end{equation}

The choice of aggregation weights $\{w_i\}$ is the central design question addressed
in this paper.
Following prior work~\cite{DBLP:journals/jsac/XieS23}, only the actor (policy)
parameters are aggregated, while critic networks remain local to each client.

\subsection{The Plateau Problem}
\label{sec:problem:plateau}

\begin{figure}[tbp]
  \centering
  \includegraphics[width=0.85\columnwidth]{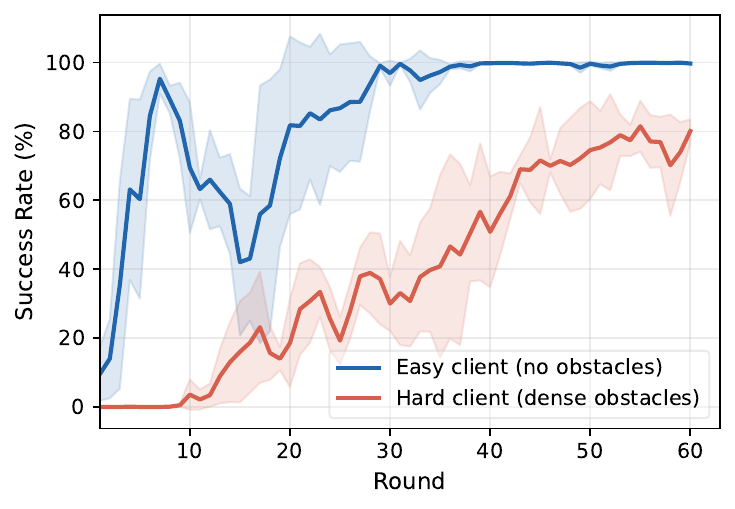}
  \caption{Training curves for one easy client (no obstacles) and one hard client
  (dense obstacles) under FedAvg~\cite{DBLP:conf/aistats/McMahanMRHA17} over three seeds.
  Shaded regions show standard deviation. The easy client's success rate saturates near
  round 30, while the hard client continues to improve throughout training, illustrating
  the plateau problem: uniform averaging weights both clients equally, even though only
  one is still contributing new information.}
  \label{fig:plateau}
\end{figure}

As training progresses, clients in easy environments converge rapidly, and their local
policies reach near-maximum performance early in training.
Once saturated, these clients continue to produce high episode returns, but their local
updates carry minimal new information. The global model already performs well in their
environment, and further local training yields negligible improvement.
We refer to this as the \emph{plateau problem}: a client has plateaued when additional
local training no longer meaningfully advances the shared policy, as seen for the easy
client in Figure~\ref{fig:plateau}.

We formally define the \emph{intra-round improvement} of client $i$ at round $k$ as
\begin{equation}
  \Delta_i^k = r_i^{\mathrm{post}} - r_i^{\mathrm{pre}},
  \label{eq:delta}
\end{equation}
where $r_i^{\mathrm{pre}}$ is the average episode return of the \emph{global} model
in $\mathcal{E}_i$ before local training, and $r_i^{\mathrm{post}}$ is the average
episode return of the \emph{local} model after training.
A plateaued client therefore satisfies $\Delta_i^k \approx 0$, indicating that
local training produced no meaningful improvement over the global model.

Under uniform averaging, plateaued easy clients contribute low-information updates yet
receive the same aggregation weight as hard clients that are still actively learning.
Under return-weighted aggregation, plateaued easy clients may receive \emph{even higher}
weights, because their absolute returns remain high after saturation.
In both cases, the aggregation signal fails to distinguish between clients
that are still improving the shared policy and those that have plateaued.
These observations motivate a weighting scheme that (i) conditions on improvement rather than absolute
return, and (ii) prioritizes harder environments among the clients that are still
improving.

\begin{figure*}[tbp]
  \centering
  \includegraphics[width=\textwidth]{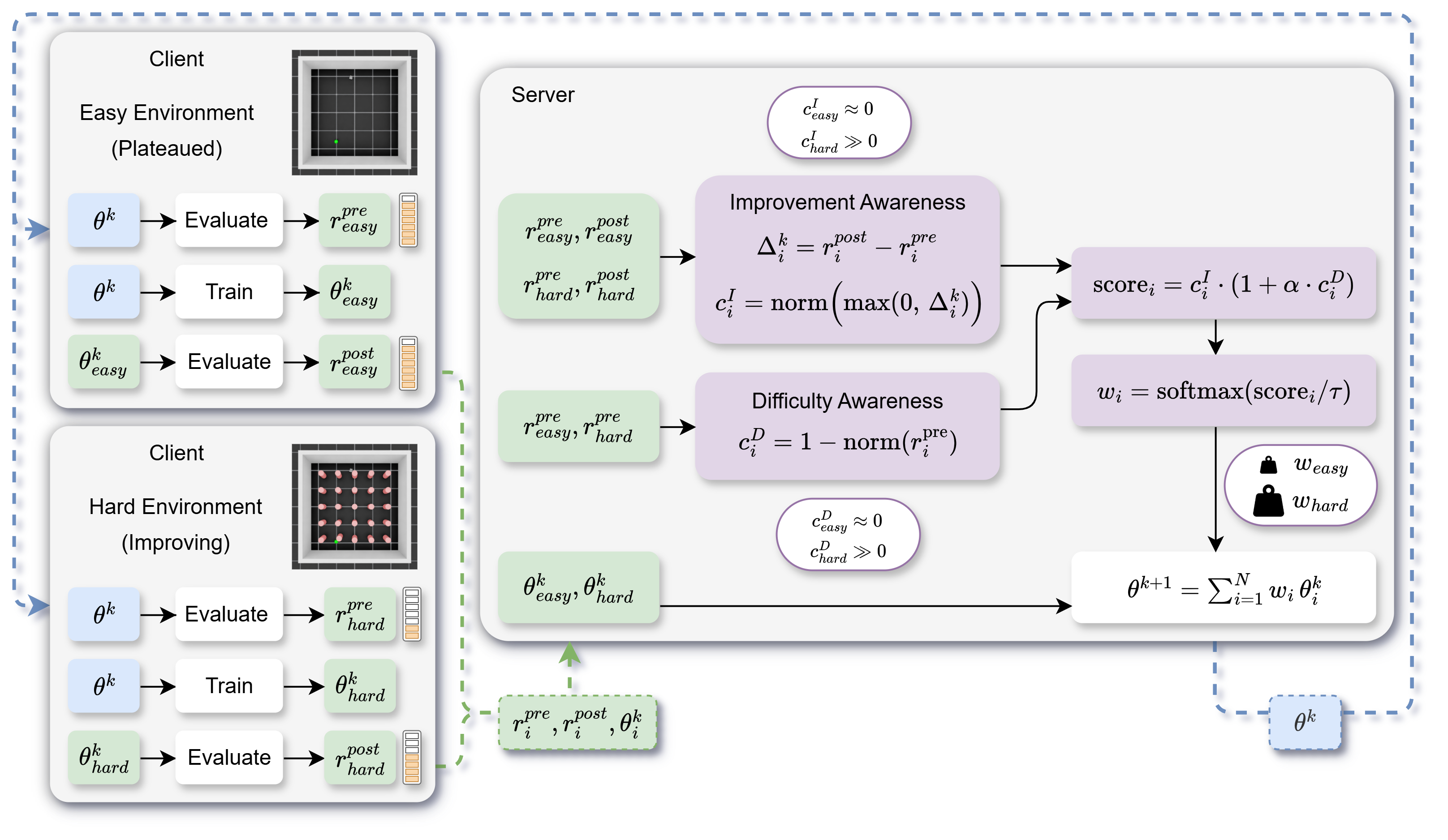}
  \caption{One communication round of FedDIA. The server broadcasts the global model
  $\theta^k$. Each client evaluates it before local training and its updated model
  after, obtaining returns $r_i^{\mathrm{pre}}$ and $r_i^{\mathrm{post}}$. Both are sent
  to the server with updated parameters $\theta_i^k$. From these, the server derives an
  improvement signal $c_i^I$ and a difficulty signal $c_i^D$, combines them into a
  per-client score, and computes aggregation weights by softmax.
  Across the two example clients, a plateaued easy client receives low weight, while a
  still-improving hard client receives high weight.}
  \label{fig:method}
\end{figure*}

\section{FedDIA}
\label{sec:method}

We propose \emph{FedDIA}, a round-level client weighting scheme that meets both
requirements.
Figure~\ref{fig:method} gives an overview of one communication round in which each
client performs evaluation before and after local training. From these returns, the
server computes difficulty- and improvement-aware aggregation signals.

\subsection{Shared Baseline}
\label{sec:method:anchor}

At the start of round $k$, each client $i$
evaluates the received global model $\theta^k$ for $E$ episodes in its local
environment $\mathcal{E}_i$, obtaining the average episode return $r_i^{\mathrm{pre}}$.
After $T$ timesteps of local training, the client evaluates the resulting model
$\theta_i^k$ for another $E$ episodes, obtaining $r_i^{\mathrm{post}}$.
The intra-round improvement $\Delta_i^k = r_i^{\mathrm{post}} - r_i^{\mathrm{pre}}$,
defined in \eqref{eq:delta}, quantifies how much local training improved upon the
global model.
Because each client's delta $\Delta_i^k$ is measured from the same baseline $\theta^k$,
their values are directly comparable across clients that train in different
environments. A higher value indicates greater improvement over the shared model and
should therefore receive a larger aggregation weight.
The shared baseline also provides a measure of environment difficulty. Since every
client obtains $r_i^{\mathrm{pre}}$ by evaluating the same global model $\theta^k$, any
difference across clients reflects the environment rather than the model. This makes
$r_i^{\mathrm{pre}}$ a natural probe of difficulty. A lower value therefore indicates a
harder environment, where the global model already performs poorly before any local
training begins.

\subsection{Contribution Signals}
\label{sec:method:signals}

We derive two normalized signals, an improvement signal $c_i^I$ from $\Delta_i^k$ and a
difficulty signal $c_i^D$ from $r_i^{\mathrm{pre}}$. Both are computed using
min-max normalization across clients, defined for any $x_i$ as:
\begin{equation}
  \norm(x_i) = \frac{x_i - \min_{j} x_j}{\max_{j} x_j - \min_{j} x_j},
  \qquad j \in \{1, \ldots, N\}.
  \label{eq:minmax}
\end{equation}

\paragraph{Improvement signal.}
A negative $\Delta_i^k$ carries no useful aggregation signal, so we clamp it to
zero before normalizing:
\begin{equation}
  c_i^I = \norm\bigl(\max(0,\, \Delta_i^k)\bigr).
  \label{eq:improv}
\end{equation}
This ensures $c_i^I = 0$ exactly for any client that fails to improve.

\paragraph{Difficulty signal.}
We map a low pre-training return to high difficulty:
\begin{equation}
  c_i^D = 1 - \norm(r_i^{\mathrm{pre}}).
  \label{eq:difficulty}
\end{equation}
This formulation requires no explicit environment metadata such as obstacle counts
or map complexity. Difficulty is measured entirely through the current capability of
the shared policy, and adapts automatically as the global model improves.

\subsection{Aggregation Rule}
\label{sec:method:score}

Simply rewarding hard-environment clients with higher weight, however, ignores whether
those clients actually made progress.
A plateaued hard-environment client might offer no more useful gradient signal than a
plateaued easy client.
We therefore treat environment difficulty not as a direct measure of contribution, but
as a factor that amplifies the score of clients that are actively improving.
This motivates the following aggregation score:
\begin{equation}
  \mathrm{score}_i =
    \underbrace{c_i^I}_{\text{improvement}} +
    \underbrace{\alpha \cdot c_i^I \cdot c_i^D}_{\text{difficulty bonus}}
    = c_i^I \cdot \bigl(1 + \alpha \cdot c_i^D\bigr),
  \label{eq:score}
\end{equation}
where $\alpha \geq 0$ is the amplification coefficient, controlling the influence of
environment difficulty on the score.
Aggregation weights are then derived from scores via a softmax with temperature
$\tau > 0$:
\begin{equation}
  w_i = \frac{\exp\bigl(\mathrm{score}_i / \tau\bigr)}
             {\sum_{j=1}^{N} \exp\bigl(\mathrm{score}_j / \tau\bigr)}.
  \label{eq:softmax}
\end{equation}

\begin{algorithm}[tbp]
\caption{FedDIA — one communication round}
\label{alg:feddia}
\begin{algorithmic}[1]
\Require Global parameters $\theta^k$; client environments $\{\mathcal{E}_i\}_{i=1}^N$;
         coefficient $\alpha$; temperature $\tau$; local timesteps $T$; episodes $E$
\Ensure  Updated global parameters $\theta^{k+1}$

\State \textit{Server:} broadcast $\theta^k$ to all clients

\For{\textit{Client} $i = 1, \ldots, N$} \textbf{in parallel}
    \State $r_i^{\mathrm{pre}} \leftarrow \mathrm{Eval}(\theta^k,\ \mathcal{E}_i,\ E)$
    \Comment{pre-eval on global model}
    \State $\theta_i^k \leftarrow \mathrm{Train}(\theta^k,\ \mathcal{E}_i,\ T)$
    \Comment{local training}
    \State $r_i^{\mathrm{post}} \leftarrow \mathrm{Eval}(\theta_i^k,\ \mathcal{E}_i,\ E)$
    \Comment{post-eval on local model}
    \State Send $r_i^{\mathrm{pre}},\ r_i^{\mathrm{post}},\ \theta_i^k$ to server
\EndFor

\State \textit{Server:} $\Delta_i^k \leftarrow r_i^{\mathrm{post}} - r_i^{\mathrm{pre}}$
       for each $i$

\If{$\max_i \Delta_i^k \leq 0$}
    \State $w_i \leftarrow 1/N$ for each $i$
    \Comment{uniform fallback}
\Else
    \State $c_i^I \leftarrow \norm(\max(0, \Delta_i^k))$
    \Comment{\eqref{eq:improv}}
    \State $c_i^D \leftarrow 1 - \norm(r_i^{\mathrm{pre}})$
    \Comment{\eqref{eq:difficulty}}
    \State $\mathrm{score}_i \leftarrow c_i^I \cdot (1 + \alpha \cdot c_i^D)$
    \Comment{\eqref{eq:score}}
    \State $w_i \leftarrow \mathrm{softmax}(\mathrm{score},\ \tau)_i$
    \Comment{\eqref{eq:softmax}}
\EndIf

\State $\theta^{k+1} \leftarrow \sum_{i=1}^{N} w_i\, \theta_i^k$
\Comment{\eqref{eq:aggregation}}

\State \Return $\theta^{k+1}$
\end{algorithmic}
\end{algorithm}

\paragraph{Improvement gate.}
If $c_i^I = 0$, then $\mathrm{score}_i = 0$, regardless of how hard the environment is.
Under the softmax, such clients receive the minimum possible weight relative to
any client that did improve, without requiring an explicit threshold.
This satisfies requirement~(i): aggregation weight is conditioned on intra-round
improvement, rather than absolute return.
When no client improves in a given round ($\Delta_i^k \leq 0$ for all $i$),
we fall back to uniform weights $w_i = 1/N$.

\paragraph{Difficulty amplification.}
When two clients improve equally, the one in the harder environment scores higher for
any $\alpha > 0$.
Min-max normalization places the easiest client at $c_{\mathrm{easy}}^D = 0$ and the
hardest at $c_{\mathrm{hard}}^D = 1$, so for equal $c^I$ their score ratio is:
\begin{equation}
  \frac{\mathrm{score}_{\mathrm{hard}}}{\mathrm{score}_{\mathrm{easy}}}
  = \frac{c^I \cdot (1 + \alpha \cdot 1)}{c^I \cdot (1 + \alpha \cdot 0)}
  = 1 + \alpha.
  \label{eq:ratio}
\end{equation}
At $\alpha = 1.0$, the hardest client therefore receives a score twice that of the
easiest.
This satisfies requirement~(ii): the preference for harder environments is built
into the score structure.
Algorithm~\ref{alg:feddia} summarizes one communication round of FedDIA.

\begin{figure*}[tbp]
  \centering
  \begin{subfigure}[b]{0.19\textwidth}
    \includegraphics[trim={120bp 120bp 120bp 120bp}, clip, width=\linewidth]{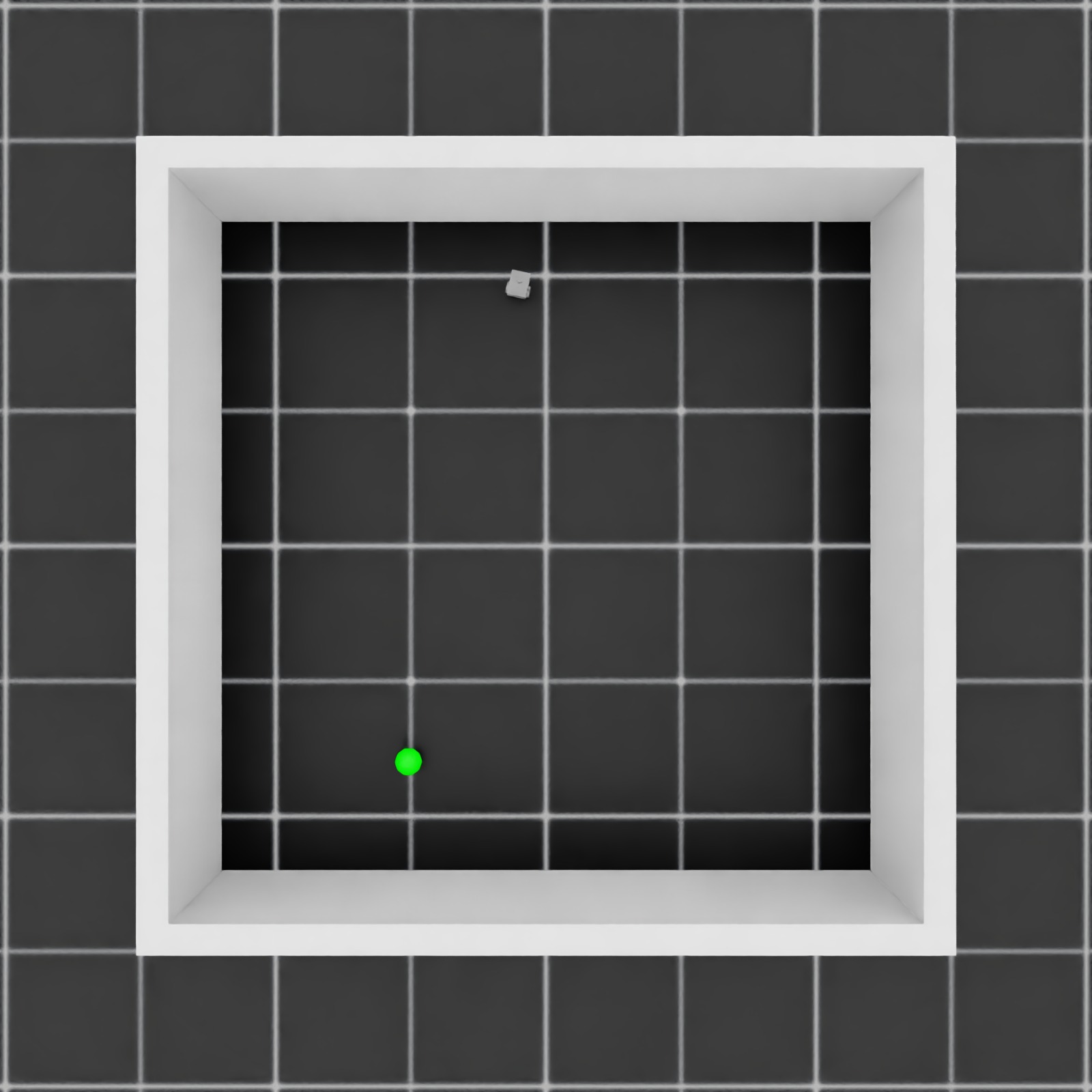}
    \caption{Empty}
    \label{fig:env:empty}
  \end{subfigure}
  \hfill
  \begin{subfigure}[b]{0.19\textwidth}
    \includegraphics[trim={120bp 120bp 120bp 120bp}, clip, width=\linewidth]{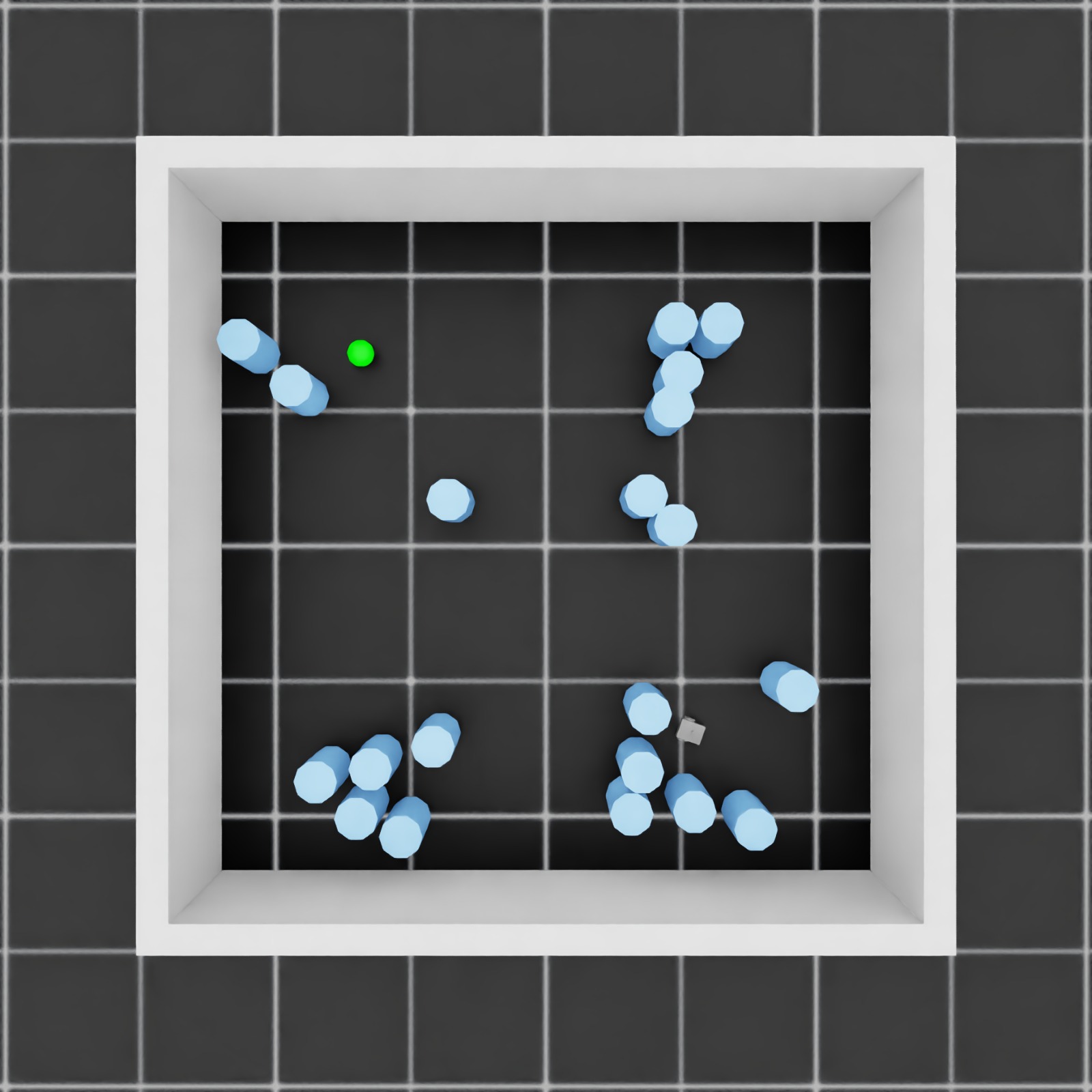}
    \caption{Random}
    \label{fig:env:random}
  \end{subfigure}
  \hfill
  \begin{subfigure}[b]{0.19\textwidth}
    \includegraphics[trim={120bp 120bp 120bp 120bp}, clip, width=\linewidth]{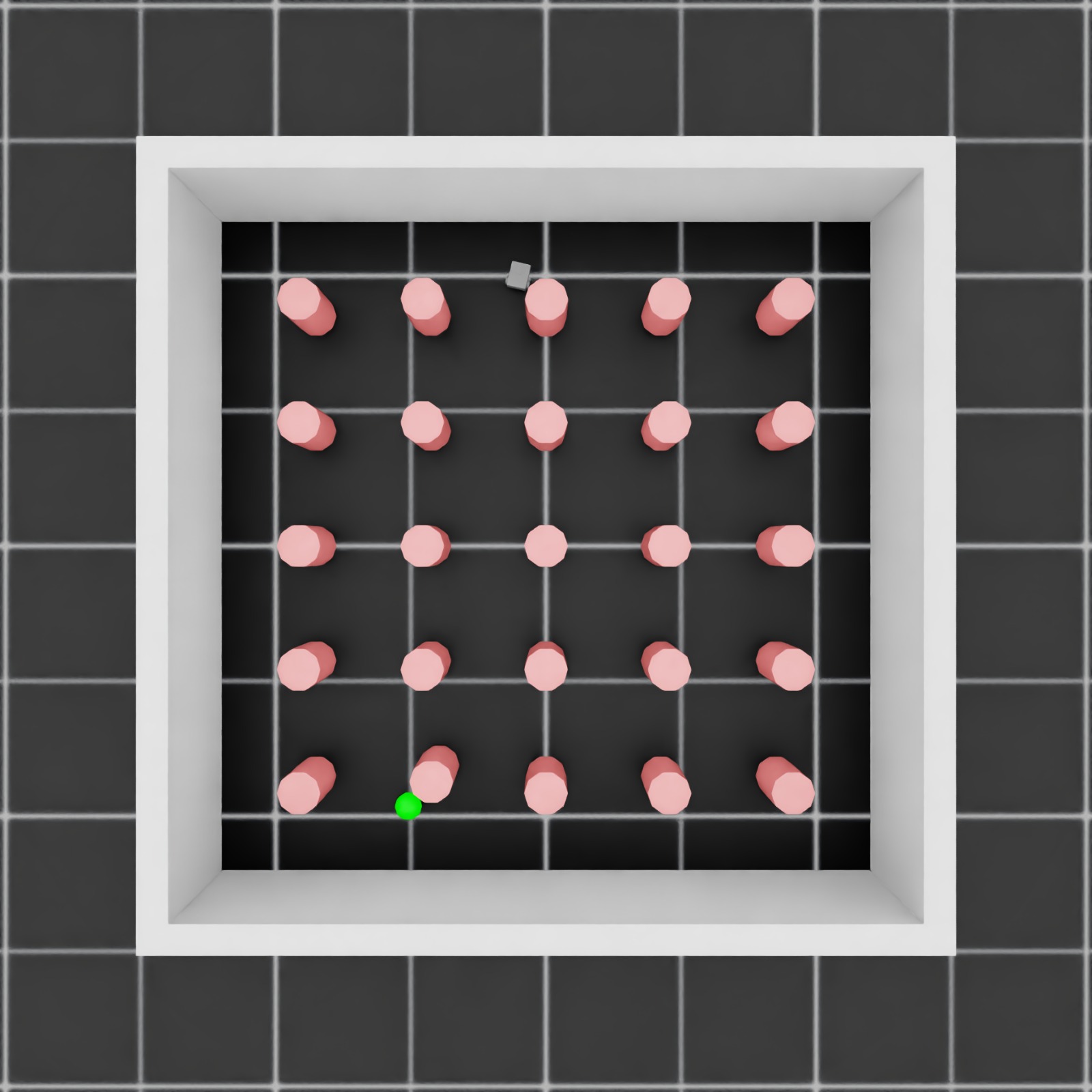}
    \caption{Grid}
    \label{fig:env:grid}
  \end{subfigure}
  \hfill
  \begin{subfigure}[b]{0.19\textwidth}
    \includegraphics[trim={120bp 120bp 120bp 120bp}, clip, width=\linewidth]{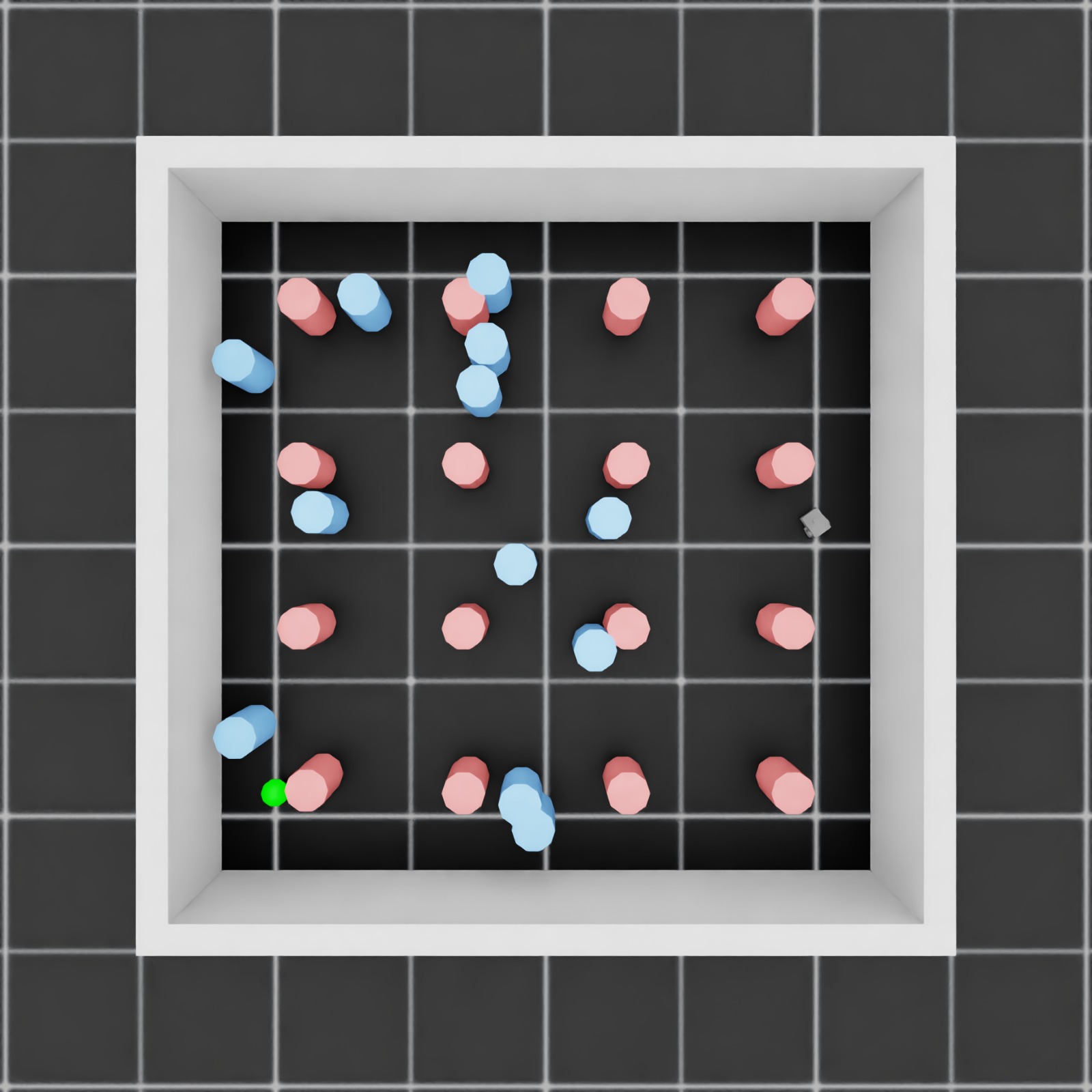}
    \caption{Mixed}
    \label{fig:env:mixed}
  \end{subfigure}
  \hfill
  \begin{subfigure}[b]{0.19\textwidth}
    \includegraphics[trim={120bp 120bp 120bp 120bp}, clip, width=\linewidth]{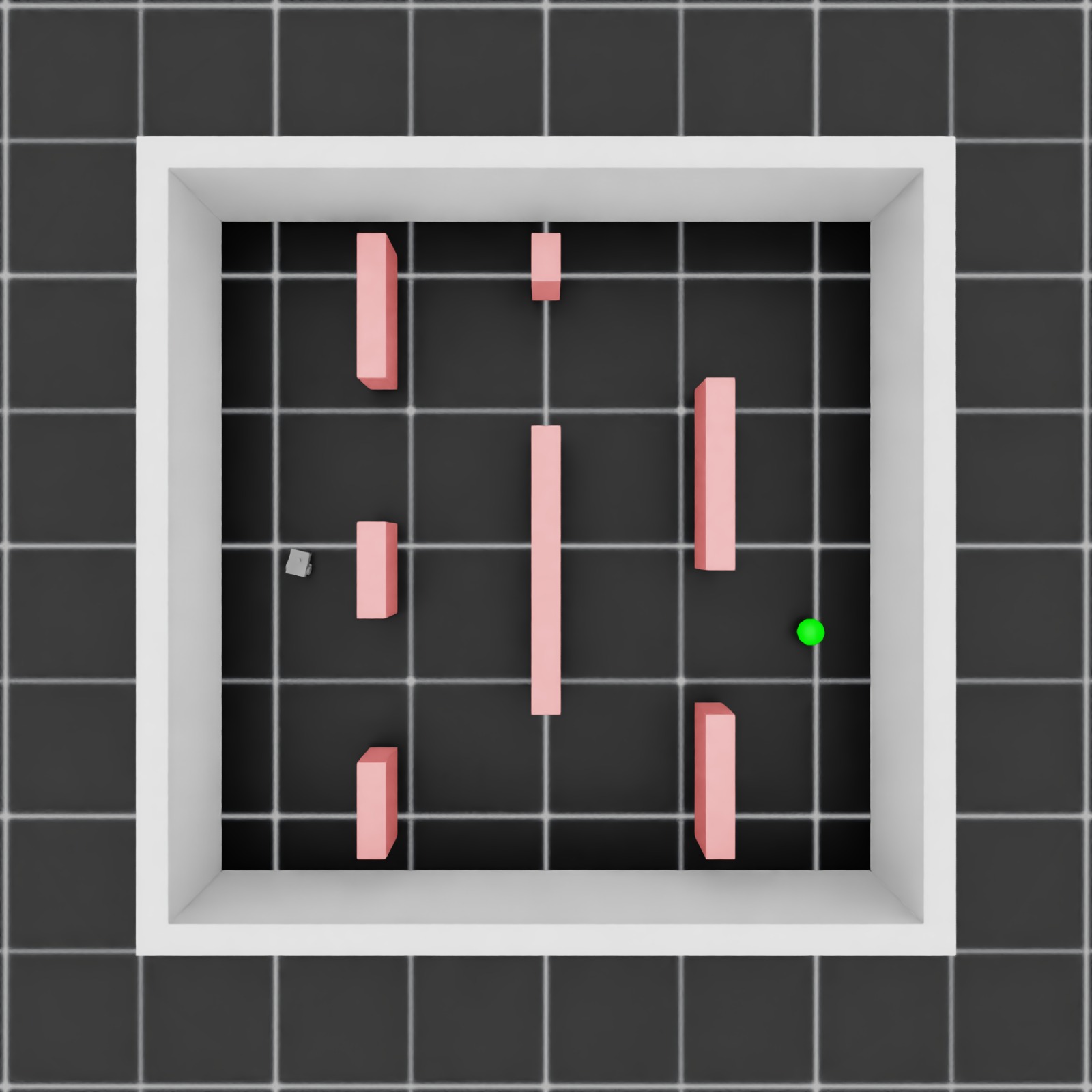}
    \caption{Multi-gate}
    \label{fig:env:multigate}
  \end{subfigure}
  \caption{Top-down view of the five obstacle modes. \emph{Empty}, \emph{random}
  and \emph{grid} are used for training, while \emph{mixed} and
  \emph{multi-gate} are held out for evaluation. Random obstacles are shown in
  blue and grid obstacles in red. The goal is the green marker.}
  \label{fig:environments}
\end{figure*}

\section{Experiments}
\label{sec:experiments}

\subsection{Experimental Setup}
\label{sec:experiments:setup}

\paragraph{Simulation Setup.}
We conduct all experiments in Isaac Lab~\cite{DBLP:journals/corr/abs-2511-04831,
DBLP:conf/nips/MakoviychukWGLS21}, a
GPU-accelerated physics simulation framework.
Each client controls a differential-drive robot navigating a
$5\,\mathrm{m} \times 5\,\mathrm{m}$ square arena enclosed by solid walls.
The task is mapless point-goal navigation in which the robot must reach a randomly placed goal
using only its current sensor readings, with no map or prior knowledge of the environment.
At the start of each episode, the robot and goal are initialized at random positions on
opposite sides of the arena, with a uniformly sampled initial heading for the robot.
The physics engine runs at 100\,Hz, and the policy is executed once every four
simulation steps, yielding a control frequency of 25\,Hz (40\,ms per decision step).
An episode ends when the robot reaches the goal (within 0.20\,m), collides with an
obstacle (minimum LiDAR reading below 0.12\,m), or exhausts the step budget of 1000
steps (40\,s).

\paragraph{Obstacle Configurations.}
Obstacles are cylindrical pillars (radius 0.15\,m, height 0.5\,m).
Three obstacle modes are used during training:
\textbf{Empty} — an open arena with no obstacles;
\textbf{Random} — 20 obstacles placed uniformly at random, re-sampled each episode;
and \textbf{Grid} — 25 obstacles in a fixed $5{\times}5$ grid, static across episodes.
Two held-out modes are reserved for evaluation:
\textbf{Mixed} — 12 randomly placed and 16 grid-placed obstacles (28 total), combining the \emph{random} and \emph{grid} configurations;
and \textbf{Multi-gate} — three parallel walls spanning the arena, each with two traversable gaps.
In all modes, obstacles are displaced at least 0.35\,m from the robot and goal spawn positions to ensure feasible episodes.
Figure~\ref{fig:environments} shows a top-down view of all five modes.

\paragraph{State Space.}
The observation vector $\mathbf{s}_t \in \mathbb{R}^{124}$ contains 120 LiDAR readings over 360° (3° resolution), each normalized by the maximum range of 5\,m with out-of-range rays set to 1.
The remaining four entries encode the goal and robot state:
\textbf{distance to goal}, the Euclidean distance to the goal normalized by the arena diagonal ($5\sqrt{2}$\,m);
\textbf{goal angle}, the angle to the goal as seen by the robot, computed as $\mathrm{atan2}(\Delta y, \Delta x) - \psi_t$ and normalized by $\pi$ after wrapping to $[-\pi, \pi]$, where $\psi_t$ is the robot's heading;
\textbf{linear velocity}, forward speed normalized by 0.22\,m/s;
and \textbf{angular velocity}, rotation rate normalized by 2.0\,rad/s.
All entries are clamped to $[-5, 5]$.

\paragraph{Action Space.}
The action space is continuous and two-dimensional, consisting of linear velocity
$v_t \in [0,\ 0.22]\,\mathrm{m/s}$ and angular velocity $\omega_t \in [-2.0,\ 2.0]\,\mathrm{rad/s}$.
The actor outputs values in $[-1, 1]$, which are linearly rescaled to each physical range.
Since linear velocity is non-negative, the robot can only move forward or stop.

\paragraph{Reward.}
The per-step reward is the sum of six terms:
\begin{equation*}
r_t = r^{\mathrm{goal}} + r^{\mathrm{col}} + r^{\mathrm{prog}} + r^{\mathrm{step}} + r^{\mathrm{fwd}} + r^{\mathrm{smooth}}.
\end{equation*}
The goal and collision terms are sparse, taking a nonzero value only when the robot
reaches the goal or collides with an obstacle:
\begin{align*}
r^{\mathrm{goal}} &= \begin{cases} 200 & \text{if } \delta_t < 0.20\,\mathrm{m}, \\ 0 & \text{otherwise,} \end{cases} \\
r^{\mathrm{col}}  &= \begin{cases} -40 & \text{if } d_{\min} < 0.12\,\mathrm{m}, \\ 0 & \text{otherwise,} \end{cases}
\end{align*}
where $\delta_t$ is the distance to goal at step $t$ and $d_{\min}$ is the minimum of the 120 LiDAR readings.
The remaining four terms are dense, applied at every step:
\begin{align*}
r^{\mathrm{prog}}   &= 10 \cdot (\delta_{t-1} - \delta_t), \\
r^{\mathrm{step}}   &= -0.05, \\
r^{\mathrm{fwd}}    &= 0.25 \cdot v_t, \\
r^{\mathrm{smooth}} &= -0.1 \cdot \lvert\omega_t - \omega_{t-1}\rvert,
\end{align*}
where $v_t$ and $\omega_t$ are the linear and angular velocities at step $t$.
$r^{\mathrm{prog}}$ rewards progress toward the goal;
$r^{\mathrm{step}}$ penalizes each timestep to encourage efficiency;
$r^{\mathrm{fwd}}$ encourages forward motion;
and $r^{\mathrm{smooth}}$ penalizes abrupt changes in angular velocity between consecutive steps to promote smooth turning.
The total reward is clipped to $[-500, 500]$.

\paragraph{Client Configuration.}
We use $N = 8$ clients, each assigned a single obstacle mode for the duration of training.
To reproduce the plateau scenario described in Section~\ref{sec:problem:plateau}, in which a
majority of clients converge early while a minority continue to offer meaningful learning
signal, we assign four clients to \emph{empty}, two to \emph{random}, and two to
\emph{grid}.

\begin{table}[tbp]
\centering
\caption{TD3 hyperparameters.}
\label{tab:hyperparams}
\begin{tabular}{ll}
\toprule
Parameter & Value \\
\midrule
Learning rate                      & $3 \times 10^{-4}$ \\
Replay buffer size                 & $10^6$ \\
Batch size                         & 1024 \\
Discount factor $\gamma$           & 0.995 \\
Exploration noise $\sigma$         & 0.3 \\
\bottomrule
\end{tabular}
\end{table}

\begin{table*}[tbp]
\centering
\caption{Success rate (\%) across training and held-out environments.
Each method is trained for 60 communication rounds and the resulting model is evaluated for 1000 episodes in every environment.
Results are reported as mean $\pm$ standard deviation across three independent seeds. Best federated result per column in bold.
\emph{Local} trains each client alone with no aggregation and is reported as a non-federated reference.}
\label{tab:main}
\resizebox{0.8\textwidth}{!}{
\begin{tabular}{lcccccc}
\toprule
 & \multicolumn{3}{c}{Training environments} & \multicolumn{2}{c}{Held-out environments} &  \\
\cmidrule(lr){2-4} \cmidrule(lr){5-6}
Method & empty & random & grid & mixed & multi-gate & Mean \\
\midrule
Local (Empty) & $85.9 \pm 21.3$ & $11.8 \pm \phantom{0}2.7$ & $\phantom{0}0.0 \pm \phantom{0}0.0$ & $\phantom{0}0.2 \pm \phantom{0}0.2$ & $\phantom{0}2.4 \pm \phantom{0}1.2$ & $20.1 \pm \phantom{0}4.6$ \\
Local (Random) & $95.6 \pm \phantom{0}1.6$ & $71.2 \pm \phantom{0}9.1$ & $61.1 \pm \phantom{0}9.8$ & $48.3 \pm 10.2$ & $67.5 \pm 12.2$ & $68.8 \pm \phantom{0}8.3$ \\
Local (Grid) & $53.9 \pm 20.3$ & $54.7 \pm 10.8$ & $41.5 \pm 24.3$ & $34.3 \pm 11.5$ & $48.3 \pm \phantom{0}7.5$ & $46.5 \pm 14.8$ \\
\midrule
FedAvg & $99.2 \pm \phantom{0}0.7$ & $73.1 \pm \phantom{0}2.1$ & $61.4 \pm 12.9$ & $49.4 \pm \phantom{0}5.3$ & $60.2 \pm \phantom{0}1.6$ & $68.7 \pm \phantom{0}4.3$ \\
Reward-Weighted & $\mathbf{99.9 \pm \phantom{0}0.1}$ & $64.1 \pm \phantom{0}3.0$ & $35.9 \pm \phantom{0}1.8$ & $33.7 \pm \phantom{0}0.6$ & $55.6 \pm \phantom{0}3.3$ & $57.9 \pm \phantom{0}1.5$ \\
FGRL & $98.9 \pm \phantom{0}0.6$ & $63.6 \pm \phantom{0}3.0$ & $32.5 \pm \phantom{0}9.0$ & $31.8 \pm \phantom{0}5.0$ & $54.8 \pm \phantom{0}7.0$ & $56.3 \pm \phantom{0}4.6$ \\
FedKL & $99.2 \pm \phantom{0}0.5$ & $72.4 \pm \phantom{0}5.4$ & $58.4 \pm 15.4$ & $45.4 \pm 12.2$ & $59.0 \pm \phantom{0}6.5$ & $66.9 \pm \phantom{0}7.9$ \\
FedProx & $99.6 \pm \phantom{0}0.3$ & $73.3 \pm \phantom{0}3.0$ & $56.6 \pm \phantom{0}5.8$ & $43.9 \pm \phantom{0}5.9$ & $64.5 \pm \phantom{0}3.1$ & $67.6 \pm \phantom{0}3.4$ \\
\midrule
FedDIA & $98.5 \pm \phantom{0}0.4$ & $\mathbf{77.4 \pm \phantom{0}5.5}$ & $\mathbf{75.1 \pm \phantom{0}7.6}$ & $\mathbf{59.7 \pm \phantom{0}3.7}$ & $\mathbf{66.8 \pm \phantom{0}7.3}$ & $\mathbf{75.5 \pm \phantom{0}3.5}$ \\
\bottomrule
\end{tabular}
}
\end{table*}

\paragraph{Local Training.}
Each client trains with TD3~\cite{DBLP:conf/icml/FujimotoHM18}, an off-policy actor-critic algorithm for
continuous action spaces.
Both the actor and critic are two-layer MLPs with 256 units per layer.
As noted in Section~\ref{sec:problem:frl}, only the actor is aggregated, and the
critic remains local throughout training.
Key TD3 hyperparameters are listed in Table~\ref{tab:hyperparams}.

\paragraph{Federated Configuration.}
We train for $K = 60$ communication rounds under three independent seeds.
At each round, every client evaluates the received global model for $E = 200$ episodes
to obtain $r_i^{\mathrm{pre}}$, then trains locally for $T = 300{,}000$ timesteps.
It then evaluates the resulting local model for another $E = 200$ episodes to obtain
$r_i^{\mathrm{post}}$.
Both are undiscounted episode returns, averaged over the $E$ evaluation episodes,
with $\gamma$ applying to local training only.
Relative to baselines that weight clients by measured performance, the additional
pre-training evaluation accounts for about $15\%$ of round wall-clock time.
FedDIA combines $r_i^{\mathrm{pre}}$ and $r_i^{\mathrm{post}}$ into an aggregation score
\eqref{eq:score} using amplification coefficient $\alpha = 1.0$, then derives weights via
softmax \eqref{eq:softmax} with temperature $\tau = 1.0$.

\paragraph{Baselines.}
We compare FedDIA against five federated baselines:
\textbf{FedAvg}~\cite{DBLP:conf/aistats/McMahanMRHA17} (uniform weights),
\textbf{Reward-Weighted} (weights proportional to average episode return),
\textbf{FGRL}~\cite{tian2024fgrl} (cross-round growth-based
weights\footnote{Following the original work, FGRL also aggregates the
critic and applies a soft local update ($\lambda = 0.9$).}),
\textbf{FedKL}~\cite{DBLP:journals/jsac/XieS23} (KL-regularized local
training\footnote{KL divergence cannot be directly computed for TD3's deterministic policy. We
penalize the squared difference of actor outputs, which is proportional to the KL
between the corresponding Gaussian exploration policies. We
sweep and fix the coefficients ($\lambda_1 = 0.1$, $\lambda_2 = 1.0$).}),
and \textbf{FedProx}~\cite{DBLP:conf/mlsys/LiSZSTS20} (proximal-regularized local
training\footnote{We use a proximal coefficient of $\mu = 1.0$.}).
We additionally report \textbf{Local}, a non-federated reference in which a client
trains independently, without aggregation.
We train one client in each of the three training environments.

\paragraph{Evaluation.}
We evaluate the resulting global model for 1000 episodes in each of the five obstacle
modes described above: the three training modes (\emph{empty}, \emph{random},
\emph{grid}) and two held-out modes (\emph{mixed} and \emph{multi-gate}).
Our primary metric is \textbf{success rate}, the percentage of episodes where the robot
reaches the goal without colliding.

\paragraph{Hardware.}
We run all experiments on a single machine equipped with an NVIDIA GeForce
RTX~3090 (24\,GB VRAM), an Intel Core i9-12900 (16 cores, 32 threads), and 32\,GB
of system memory.
Training and evaluation both run inside a Docker container based on the Isaac Lab
image.

\subsection{Main Comparison}
\label{sec:experiments:main}

Table~\ref{tab:main} reports success rate across all five environments.
FedDIA attains the highest mean success rate ($75.5\%$ against $68.7\%$ for FedAvg,
the strongest federated baseline).
Its margin is largest in \emph{grid}, the hardest of the three training
environments, and in \emph{mixed}, the harder of the two held-out environments.
In \emph{empty}, every federated method is near ceiling and the differences
between them are negligible.

\begin{figure}[tbp]
  \centering
  \includegraphics[width=\linewidth]{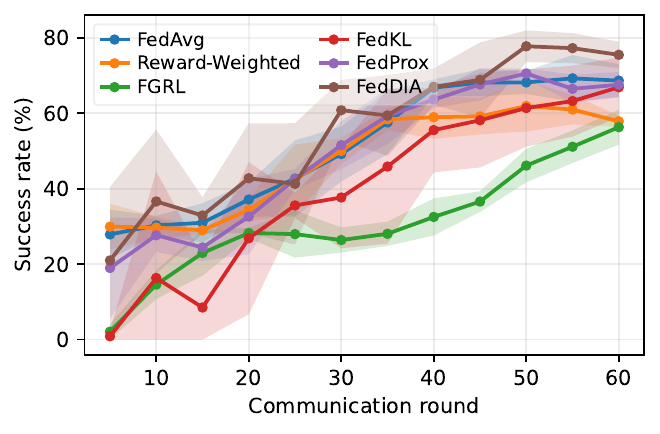}
  \caption{Success rate (\%) of the global model over communication rounds, averaged
  across the five environments (1000 episodes each). Shaded regions show one
  standard deviation over three seeds.}
  \label{fig:main_curves}
\end{figure}

Figure~\ref{fig:main_curves} tracks the same comparison across communication rounds.
FedDIA stays close to the baselines through the early rounds, and its margin widens
toward the end.
Weighting on the wrong signal can cost more than simply averaging: reward-weighted
aggregation performs worse than FedAvg (mean success rate of $57.9\%$ against
$68.7\%$).
Easy environments permit high absolute returns by construction, so weighting by
return favors the easy-environment clients and pulls the global model toward
updates that carry little new information.

The \emph{Local} rows show the limits of training alone, except
\emph{Local (Random)}, which reaches $68.8\%$ and matches FedAvg.
Re-sampling its obstacles every episode gives it both an easy layout to start
from and enough variety to perform well across our five evaluation environments.
We attribute FedAvg's result to the plateau of Section~\ref{sec:problem:plateau}
rather than to a limitation of federation itself.
Four of the eight clients train in \emph{empty}, so uniform weighting assigns
half of the aggregate to clients that have already converged, diluting the
contribution of those still improving.
FedDIA reweights toward those clients and gains $6.8$ points over FedAvg.

\subsection{Aggregation Weight Allocation}
\label{sec:experiments:weights}

\begin{figure*}[tbp]
  \centering
  \includegraphics[width=\textwidth]{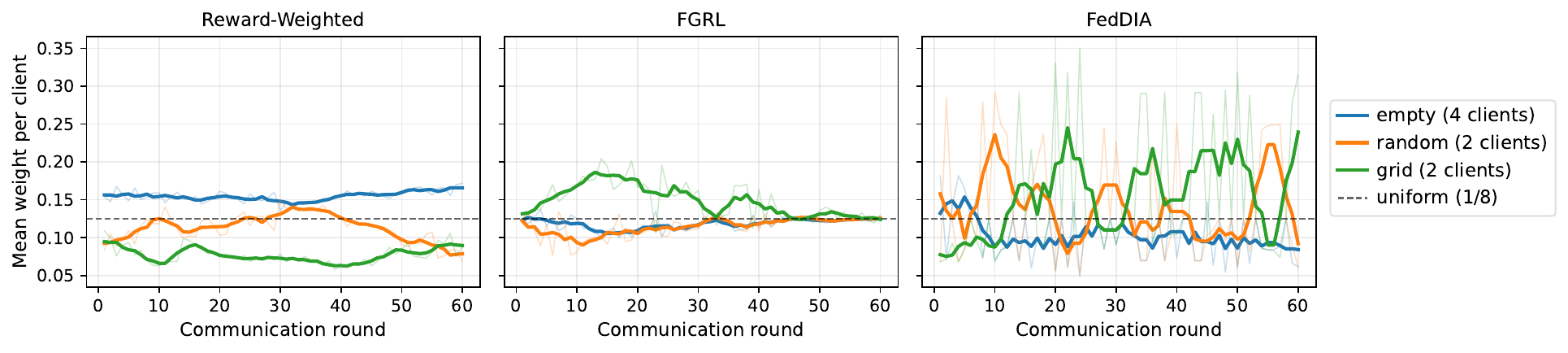}
  \caption{Aggregation weight received by a client in each environment group
  (averaged within the group) for one representative seed. The lines show a
  five-round moving average. The dashed line at $1/8 = 0.125$ is the uniform
  weighting of FedAvg.}
  \label{fig:weights}
\end{figure*}

Figure~\ref{fig:weights} shows the average weight received by a client in each
environment group during training.
FedKL and FedProx are omitted since their weights are uniform, as in FedAvg.
Reward-weighted aggregation favors the \emph{empty} clients throughout, holding
their weight above the uniform level for the entire run, while the \emph{grid}
clients never rise above it.
Absolute return is highest where the task is easiest, so weighting by return
concentrates the global model on the environments that needed the least help.
FGRL's self-referential growth shrinks as training proceeds, and the separation
between groups decays until all three settle at the uniform level.

FedDIA makes improvements comparable across clients, and these improvements can
therefore be rescaled each round, allowing the spread between clients to survive
even as the improvements themselves shrink.
Under FedDIA, the \emph{empty} clients receive above-uniform weight over the
first few rounds, while the global model is still improving in every
environment. They fall below uniform thereafter, as their improvements
become smaller.
The weight allocation in FedDIA is also considerably more dynamic than either
baseline, responding to which clients improved in a given round.
Relative to the baselines, FedDIA moves weight toward the harder environments and
away from the easiest.

\begin{table}[tbp]
\centering
\caption{Mean success rate (\%) over the five evaluation environments for
amplification coefficient $\alpha$. We use a smaller experiment
setup with four clients (three \emph{empty}, one \emph{grid}) trained for 30
communication rounds. All other settings are the same as in
Section~\ref{sec:experiments:setup}. \emph{Improvement only} ($\alpha = 0$)
reduces the score to the improvement signal alone; \emph{Difficulty only}
replaces \eqref{eq:score} with $\mathrm{score}_i = c_i^D$. FedDIA uses
$\alpha = 1.0$ in all other experiments.}
\label{tab:alpha}
\resizebox{0.85\columnwidth}{!}{
\begin{tabular}{lcc}
\toprule
Method & $\alpha$ & Mean success rate \\
\midrule
Improvement only & 0.00 & $29.6 \pm \phantom{0}7.5$ \\
\midrule
\multirow{6}{*}{FedDIA} & 0.25 & $37.2 \pm 13.1$ \\
 & 0.50 & $31.3 \pm \phantom{0}4.4$ \\
 & 0.75 & $31.1 \pm \phantom{0}5.3$ \\
 & 1.00 & $36.4 \pm \phantom{0}8.4$ \\
 & 1.25 & $22.4 \pm \phantom{0}2.4$ \\
 & 1.50 & $33.8 \pm 10.9$ \\
\midrule
Difficulty only & -- & $30.6 \pm \phantom{0}5.2$ \\
\bottomrule
\end{tabular}
}
\end{table}

\subsection{Score Ablation}
\label{sec:experiments:alpha}

Table~\ref{tab:alpha} sweeps the amplification coefficient $\alpha$ of
\eqref{eq:score} and isolates each of the two signals in turn.
We use a smaller experiment setup with four clients rather than eight, three in
\emph{empty} and one in \emph{grid}, trained for 30 communication rounds instead
of 60, with every other setting as in Section~\ref{sec:experiments:setup}.
\emph{Improvement only} ($\alpha = 0$) removes the difficulty bonus and weights
clients by intra-round improvement alone, reaching $29.6\%$.
\emph{Difficulty only} ($\mathrm{score}_i = c_i^D$) removes the improvement
signal instead, weighting clients by environment difficulty whether or not they
are learning, reaching $30.6\%$.
Combining the two signals gives a higher mean success rate than either alone for
five of the six $\alpha$ values, including the default $\alpha = 1.0$ at $36.4\%$.

\begin{table*}[tbp]
\centering
\caption{Success, collision and timeout rates (\%) in the two hand-authored layouts of
Figure~\ref{fig:realistic_environments}. The final models of
Section~\ref{sec:experiments:main} are evaluated without further training, under
the same evaluation protocol. Best federated result per column in bold.
\emph{Local} is reported as a non-federated reference.}
\label{tab:realistic}
\resizebox{0.8\textwidth}{!}{
\begin{tabular}{lcccccc}
\toprule
 & \multicolumn{3}{c}{Warehouse} & \multicolumn{3}{c}{Living space} \\
\cmidrule(lr){2-4} \cmidrule(lr){5-7}
Method & success~$\uparrow$ & collision~$\downarrow$ & timeout~$\downarrow$ & success~$\uparrow$ & collision~$\downarrow$ & timeout~$\downarrow$ \\
\midrule
Local (Empty) & $\phantom{0}1.8 \pm \phantom{0}0.6$ & $68.3 \pm 50.8$ & $29.9 \pm 51.2$ & $\phantom{0}3.8 \pm \phantom{0}2.0$ & $65.0 \pm 51.4$ & $31.1 \pm 53.3$ \\
Local (Random) & $38.3 \pm \phantom{0}7.6$ & $14.5 \pm \phantom{0}8.4$ & $47.2 \pm 11.2$ & $37.1 \pm 11.0$ & $\phantom{0}7.0 \pm \phantom{0}3.4$ & $56.0 \pm \phantom{0}9.2$ \\
Local (Grid) & $33.0 \pm 15.5$ & $11.2 \pm \phantom{0}1.9$ & $55.8 \pm 14.9$ & $22.6 \pm \phantom{0}9.6$ & $\phantom{0}1.7 \pm \phantom{0}0.4$ & $75.7 \pm \phantom{0}9.2$ \\
\midrule
FedAvg & $49.4 \pm \phantom{0}4.8$ & $32.2 \pm \phantom{0}3.0$ & $18.5 \pm \phantom{0}2.4$ & $61.3 \pm \phantom{0}3.5$ & $12.9 \pm \phantom{0}4.7$ & $25.8 \pm \phantom{0}1.8$ \\
Reward-Weighted & $36.1 \pm \phantom{0}2.1$ & $53.0 \pm \phantom{0}8.6$ & $\mathbf{10.9 \pm \phantom{0}9.6}$ & $52.7 \pm \phantom{0}7.0$ & $35.8 \pm 12.8$ & $\mathbf{11.6 \pm \phantom{0}8.3}$ \\
FGRL & $28.7 \pm \phantom{0}6.9$ & $23.2 \pm 10.4$ & $48.1 \pm 17.2$ & $32.5 \pm 11.1$ & $\mathbf{\phantom{0}5.2 \pm \phantom{0}2.7}$ & $62.4 \pm 13.9$ \\
FedKL & $39.7 \pm 23.8$ & $\mathbf{16.2 \pm 11.5}$ & $44.1 \pm 35.3$ & $43.6 \pm 27.3$ & $\phantom{0}6.1 \pm \phantom{0}6.3$ & $50.3 \pm 33.2$ \\
FedProx & $50.5 \pm \phantom{0}5.0$ & $33.4 \pm 14.0$ & $16.2 \pm 11.6$ & $61.1 \pm 11.4$ & $14.8 \pm \phantom{0}9.8$ & $24.1 \pm 21.3$ \\
\midrule
FedDIA & $\mathbf{61.6 \pm \phantom{0}3.6}$ & $20.6 \pm \phantom{0}5.0$ & $17.8 \pm \phantom{0}1.7$ & $\mathbf{74.8 \pm \phantom{0}6.7}$ & $\phantom{0}6.4 \pm \phantom{0}2.1$ & $18.9 \pm \phantom{0}6.7$ \\
\bottomrule
\end{tabular}
}
\end{table*}

\begin{figure}[tbp]
  \centering
  \begin{subfigure}[b]{0.6\columnwidth}
    \includegraphics[trim={140bp 230bp 140bp 330bp}, clip, width=\linewidth]{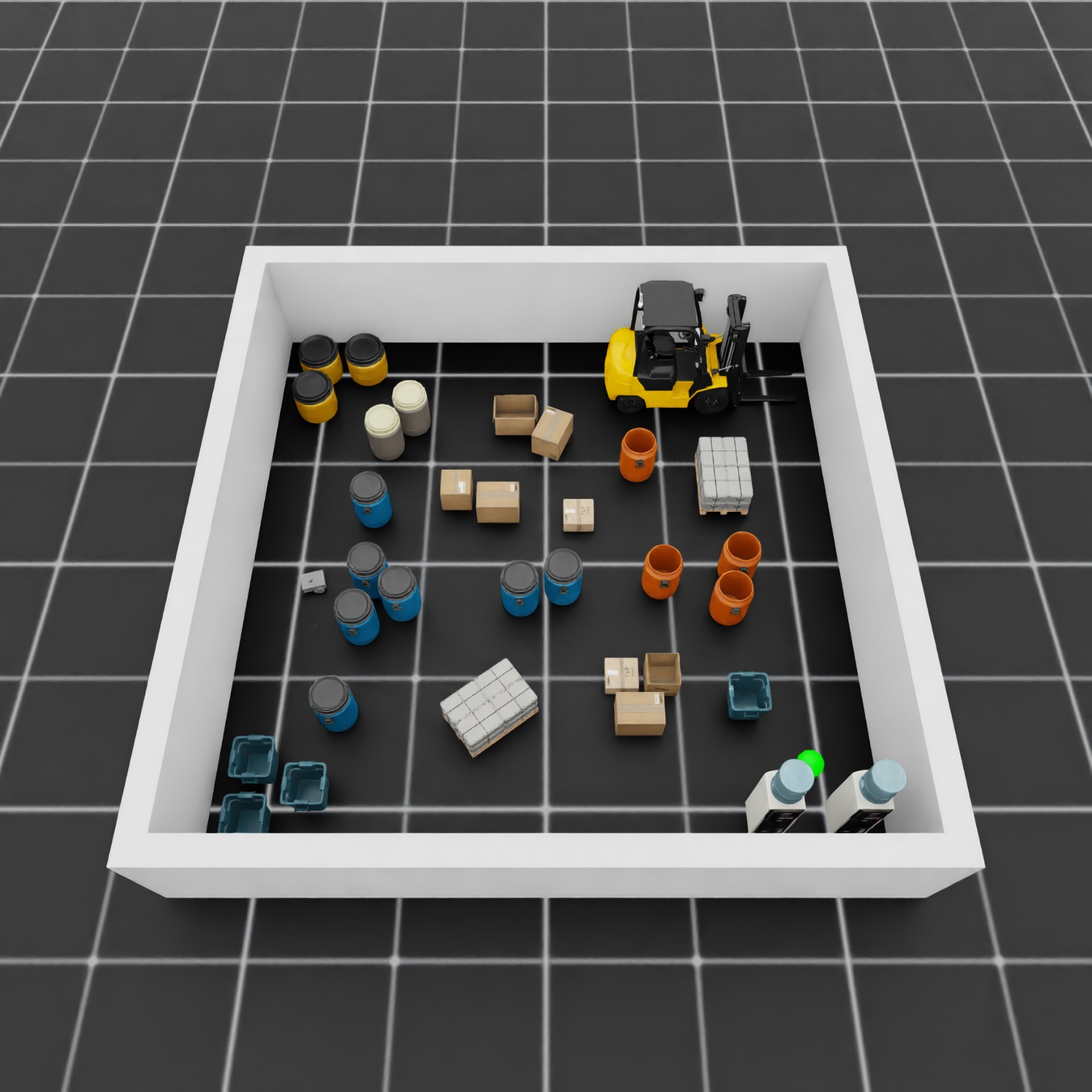}
    \caption{Warehouse}
    \label{fig:env:warehouse}
  \end{subfigure}
  \hfill
  \begin{subfigure}[b]{0.6\columnwidth}
    \includegraphics[trim={140bp 230bp 140bp 330bp}, clip, width=\linewidth]{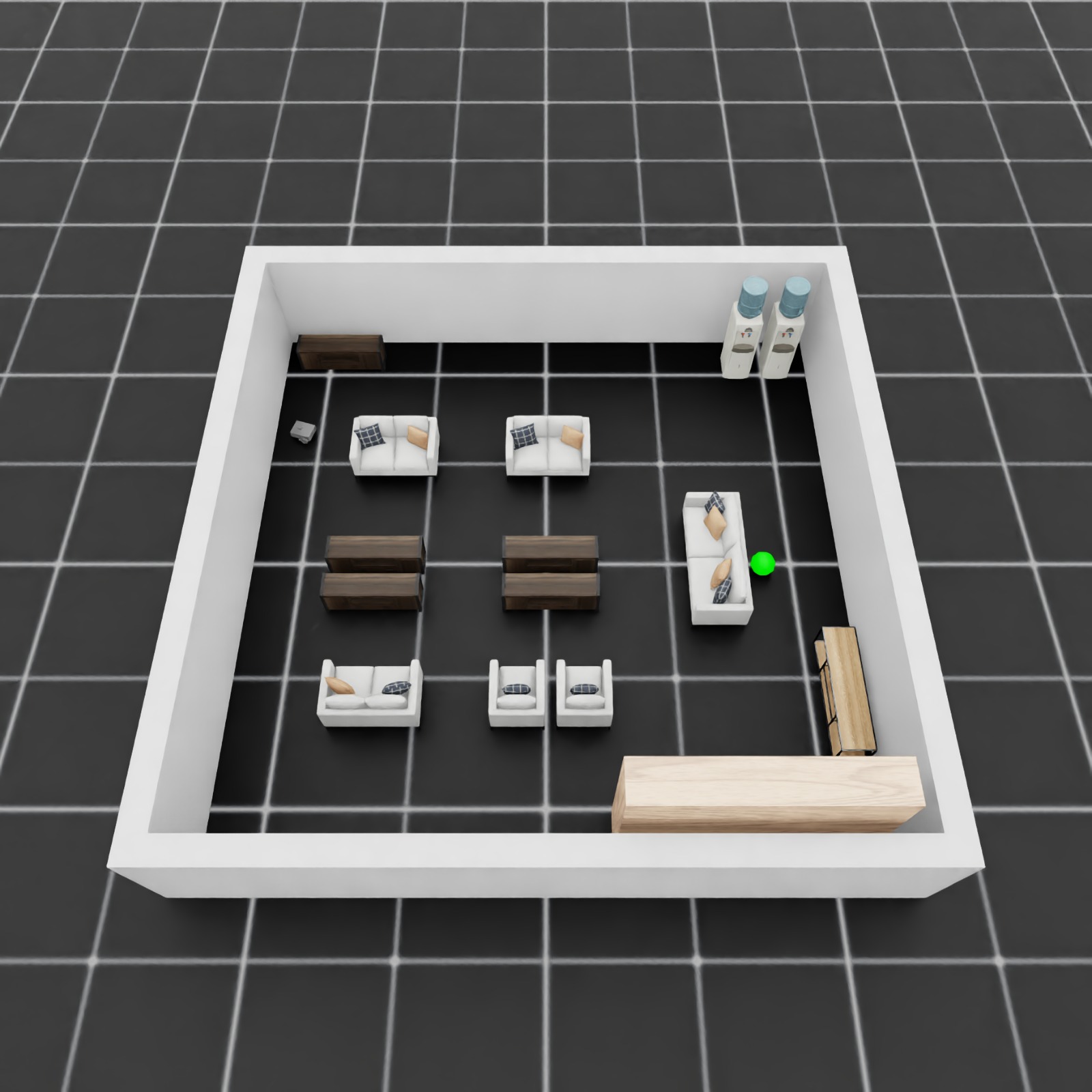}
    \caption{Living space}
    \label{fig:env:living}
  \end{subfigure}
  \caption{Hand-authored layouts, used for evaluation only: (a) a warehouse
  assembled from industrial equipment, and (b) a living space assembled from
  household furniture. The objects are fixed in both layouts and only the
  robot and goal spawn positions are re-sampled between episodes.}
  \label{fig:realistic_environments}
\end{figure}

\subsection{Structured Obstacle Layouts}
\label{sec:experiments:realistic}

To test whether policies trained only against cylindrical pillars of a single
radius transfer to obstacles of more realistic shape and varied size, we
hand-author two layouts: a warehouse of industrial equipment and a living space
of household furniture, both shown in Figure~\ref{fig:realistic_environments}.
We evaluate the same final models reported in
Section~\ref{sec:experiments:main}, without any further training.
Table~\ref{tab:realistic} reports success, collision and timeout rates for each
method in both layouts.
FedDIA reaches the highest success rate in both, $61.6\%$ against FedProx's $50.5\%$
in the warehouse and $74.8\%$ against FedAvg's $61.3\%$ in the living space.
\emph{Local (Random)}, which matched FedAvg in Table~\ref{tab:main}, reaches
only $38.3\%$ and $37.1\%$ here, against FedAvg's $49.4\%$ and $61.3\%$.
It did not transfer to unfamiliar obstacle shapes as well as FedAvg.

FedDIA's collision and timeout rates, however, are not the lowest in the table.
This is because a policy can eliminate collisions by refusing to move, and
timeouts by moving without regard for obstacles.
Reward-Weighted, for example, has the fewest timeouts ($10.9\%$ and $11.6\%$) and the
highest collision rate among the federated baselines ($53.0\%$ and $35.8\%$),
trading one failure mode for the other.
FedDIA avoids both extremes.

\subsection{Robustness to a Defective Client}
\label{sec:experiments:defective}

A deployed fleet cannot assume every robot is working properly. For example, a
drive motor may fail on one wheel, so that the commanded action is no longer the
motion the robot executes.
The model that client produces therefore drifts away from any useful policy.
The client keeps training and reporting, and its update is aggregated along with
every other client's.
We simulate such a failure by replacing one client's actions with uniform random
values during training.
We use the default configuration of
Section~\ref{sec:experiments:setup} and make one of the two \emph{grid} clients
defective.
In a federated setting the server does not know which client is defective, so any
down-weighting has to emerge from the aggregation rule itself.
Table~\ref{tab:defective} reports success rate with the defective client present.
FedDIA maintains the highest performance at $75.6\%$.
Reward-weighted aggregation discounts the defective client automatically by
weighting on return, and therefore does better than FedAvg's uniform weighting
in this setting.

\begin{table*}[tbp]
\centering
\caption{Success rate (\%) with one defective client. One of the two \emph{grid}
clients has its actions replaced with uniform random values during training.
Training and evaluation otherwise follow the clean setting of
Table~\ref{tab:main}.
Best result per column in bold.}
\label{tab:defective}
\resizebox{0.8\textwidth}{!}{
\begin{tabular}{lcccccc}
\toprule
 & \multicolumn{3}{c}{Training environments} & \multicolumn{2}{c}{Held-out environments} &  \\
\cmidrule(lr){2-4} \cmidrule(lr){5-6}
Method & empty & random & grid & mixed & multi-gate & Mean \\
\midrule
FedAvg & $99.0 \pm \phantom{0}0.6$ & $60.5 \pm \phantom{0}4.6$ & $33.6 \pm 13.7$ & $33.3 \pm \phantom{0}5.1$ & $48.8 \pm 10.1$ & $55.0 \pm \phantom{0}6.3$ \\
Reward-Weighted & $\mathbf{99.5 \pm \phantom{0}0.5}$ & $71.4 \pm \phantom{0}2.9$ & $56.6 \pm \phantom{0}4.6$ & $45.5 \pm \phantom{0}3.5$ & $61.9 \pm \phantom{0}6.4$ & $67.0 \pm \phantom{0}2.8$ \\
FGRL & $74.2 \pm 17.8$ & $43.3 \pm \phantom{0}3.2$ & $14.5 \pm \phantom{0}1.9$ & $18.0 \pm \phantom{0}2.2$ & $34.8 \pm \phantom{0}6.4$ & $36.9 \pm \phantom{0}1.8$ \\
FedKL & $99.4 \pm \phantom{0}0.3$ & $69.7 \pm \phantom{0}5.5$ & $43.2 \pm 13.2$ & $38.3 \pm \phantom{0}9.4$ & $53.5 \pm 10.5$ & $60.8 \pm \phantom{0}7.7$ \\
FedProx & $99.2 \pm \phantom{0}0.9$ & $69.0 \pm \phantom{0}2.4$ & $54.0 \pm \phantom{0}7.4$ & $41.9 \pm \phantom{0}4.2$ & $59.2 \pm \phantom{0}0.4$ & $64.7 \pm \phantom{0}2.8$ \\
\midrule
FedDIA & $99.3 \pm \phantom{0}0.7$ & $\mathbf{78.8 \pm \phantom{0}1.1}$ & $\mathbf{75.9 \pm \phantom{0}4.4}$ & $\mathbf{56.5 \pm \phantom{0}2.5}$ & $\mathbf{67.5 \pm \phantom{0}4.0}$ & $\mathbf{75.6 \pm \phantom{0}0.6}$ \\
\bottomrule
\end{tabular}
}
\end{table*}

\begin{figure}[tbp]
  \centering
  \includegraphics[width=0.95\linewidth]{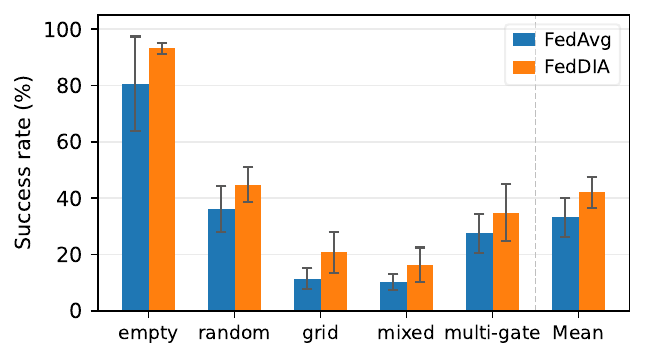}
  \caption{Success rate (\%) with $40$ clients, $8$ sampled per round, at round $60$.
  Mean $\pm$ standard deviation over three seeds.}
  \label{fig:partial}
\end{figure}

\subsection{Partial Client Participation}
\label{sec:experiments:partial}

In a deployed fleet, not every client is available in every round. We therefore
scale the number of clients to $40$ ($20$ \emph{empty}, $10$ \emph{random}, and
$10$ \emph{grid}) and sample $8$ clients uniformly at random in each round. All other
settings are unchanged.
Figure~\ref{fig:partial} reports the result. FedDIA achieves a mean success rate of
$42.0\%$, compared with $33.2\%$ for FedAvg, and leads in
each environment.

\section{Conclusion}
\label{sec:conclusion}

We identified the \emph{plateau problem} in heterogeneous federated reinforcement
learning: once a client has mastered its environment, its updates carry little new
information.
Uniform averaging nonetheless keeps weighting it equally, and return-based
weighting rewards it further.
FedDIA scores each client by how much its local training improved the global model,
amplified by the difficulty of its environment.
Both signals are measured against the same global model, so they are comparable
across clients, and require no description of the environment.
Against baselines that also weight clients by measured performance, the extra cost
is the evaluation of the global model before local training.
FedDIA reaches a mean success rate of $75.5\%$ over five evaluation environments,
against $68.7\%$ for the strongest federated baseline.
Its advantage carries over to hand-authored warehouse and living-space layouts,
evaluated without further training.
FedDIA also holds its performance when a client is defective, and
remains the best-performing method.
The plateau problem can arise wherever federated clients face environments of
varying difficulty.
In future work, we plan to evaluate FedDIA on tasks where client difficulty comes
from something other than obstacle layout.

\bibliographystyle{elsarticle-num}
\bibliography{main}

\end{document}